\documentclass[letterpaper]{article} 
\usepackage{aaai24}  
\usepackage{times}  
\usepackage{helvet}  
\usepackage{courier}  
\usepackage[hyphens]{url}  
\usepackage{graphicx} 
\usepackage{natbib}  
\usepackage{caption} 
\usepackage{algorithm}
\usepackage{algorithmic}

\usepackage{epsfig}
\usepackage{graphicx}
\usepackage{amsmath}
\usepackage{amssymb}

\usepackage{pifont}
\usepackage{bbding}
\usepackage{booktabs}
\usepackage{enumitem}
\usepackage{xcolor}
\usepackage{makecell}

\usepackage{multirow}
\usepackage{multicol}

\usepackage{newfloat}
\usepackage{listings}
\DeclareCaptionStyle{ruled}{labelfont=normalfont,labelsep=colon,strut=off} 
\floatstyle{ruled}
\newfloat{listing}{tb}{lst}{}
\floatname{listing}{Listing}
\title{MuLTI: Efficient Video-and-Language Understanding with Text-Guided MultiWay-Sampler and Multiple Choice Modeling}
\author{
    Jiaqi Xu,
    Bo Liu,
    Yunkuo Chen,
    Mengli Cheng,
    Xing Shi
}
\affiliations{
    Alibaba Group, China\\

   \{zhoumo.xjq, xuanyuan.lb, chenyunkuo.cyk, mengli.cml, shubao.sx\}@alibaba-inc.com
}

\usepackage{bibentry}

\begin{document}

\maketitle
\begin{abstract}

Video-and-language understanding has a variety of applications in the industry, such as video question answering, text-video retrieval, and multi-label classification.
Existing video-and-language understanding methods generally adopt heavy multi-modal encoders and feature fusion modules, which consume high computational costs. 
Specially, they have difficulty dealing with dense video frames or long text prevalent in industrial applications. 

This paper proposes MuLTI, a highly accurate and efficient video-and-language understanding model that achieves efficient and effective feature fusion and rapid adaptation to downstream tasks.
Specifically, we design a Text-Guided MultiWay-Sampler based on adapt-pooling residual mapping and self-attention modules to sample long sequences and fuse multi-modal features, which reduces the computational costs and addresses performance degradation caused by previous samplers.
Therefore, MuLTI can handle longer sequences with limited computational costs.
Then, to further enhance the model's performance and fill in the lack of pretraining tasks in the video question answering, we propose a new pretraining task named Multiple Choice Modeling. This task bridges the gap between pretraining and downstream tasks and improves the model's ability to align video and text features.
Benefiting from the efficient feature fusion module and the new pretraining task, MuLTI achieves state-of-the-art performance on multiple datasets.
Implementation and pretrained models will be released.
\end{abstract}
\section{Introduction}\label{sec:intro}
\begin{figure*}
    \centering 
    \includegraphics[width=0.94\textwidth]{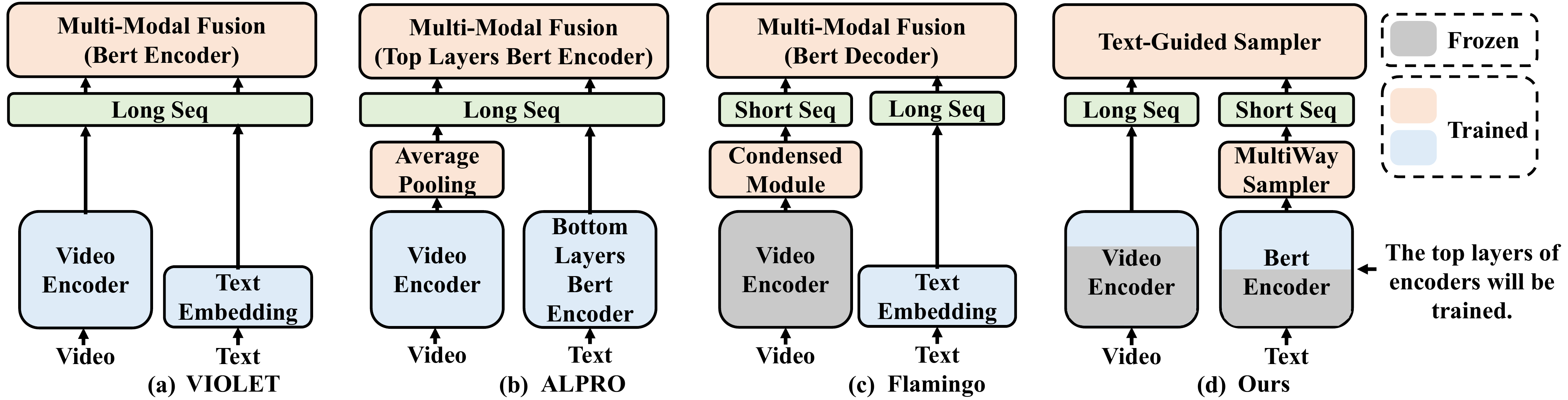}
    \caption{
        Comparison of different models. Previous works such as (a) and (b) cannot easily handle long sequences. Previous works such as (c) use randomly initialized query vectors for sampler and condense video features, which is sub-optimal solution.
    }
    \label{fig:intro}
\end{figure*}

Video-and-language understanding has a wide range of applications such as {\em video question answering~(videoQA)}, {\em text-video retrieval} and {\em multi-label classification}~\cite{Diba2019LargeSH}. 
Existing methods have made significant progress in video-and-language understanding. However, they still suffer from two challenges: 
Balancing computational efficiency and performance when dealing with long sequences and the domain gap between pretraining and downstream tasks.

The video-text model generally consists of three modules: text encoder, video encoder, and feature fusion module. The latter two usually cause high computational costs.

%
%
Feature fusion modules face efficiency and effectiveness challenges. 
Previous studies~\cite{Fu2021VIOLETE, Huang2022CloverTA} concatenate video-text encoder outputs for transformer encoders processing, with complexity growing with sequence length squared.
Other studies~\cite{Lei2021Less,Li2021AlignAP,Yang2022ZeroShotVQ,Lei2021UnderstandingCV} reduce computation by condensing video features via mean pooling or class tokens before feature fusion, risking loss of critical details.
%
%
Flamingo~\cite{Alayrac2022FlamingoAV} employs samplers and random queries for efficient video feature condensation, though this approach is suboptimal and may compromise feature integrity.
In summary, balancing computational costs and the model's accuracy in the feature fusion module is still challenging. 
Following \cite{Miech2019HowTo100MLA,Sun2019VideoBERTAJ,Li2020HeroHE,Zhu2020ActBERTLG,Miech2020EndtoEndLO}, we explore strategies for selectively freezing encoder components to lower visual encoder training costs.
%

Aligning pretraining with downstream tasks is challenging. 
Previous pretraining frameworks generally apply four typical pretraining tasks: Masked Frame Modeling (MVM) tasks~\cite{Lei2021UnderstandingCV, Ma2021Top1SO, Fu2021VIOLETE,Huang2022CloverTA} for video encoder optimization, Masked Language Modeling (MLM) tasks~\cite{Devlin2018Bert,Sun2019VideoBERTAJ,Zhu2020ActBERTLG,Luo2020Univl,Li2020HeroHE,Lei2021Less} for text encoder optimization, Video Text Matching (VTM) and Video Text Comparison (VTC) tasks~\cite{Li2020HeroHE,Luo2020Univl,Fu2021VIOLETE,Li2021AlignAP} for joint optimization of video and text encoders. 
Although the above methods have proven effective in learning video and text representations, there are still significant domain gaps between pretraining and downstream tasks, especially in videoQA. 
Only the VTC task is consistent with text-video retrieval among the above pretraining tasks.
In summary, narrowing the domain gap between the pretraining and downstream tasks is still challenging.

Addressing these challenges, we introduce MuLTI, featuring a Text-Guided MultiWay-Sampler for sequence condensation and multi-modal fusion.
Existing methods typically use a learnable query vector to sample the video feature through self-attention modules \cite{Alayrac2022FlamingoAV}. 
A randomly initialized query vector can discard vital original feature information, causing performance drops. 
We design an lightweight Adapt-Pooling method in Text-Guided MultiWay-Sampler to obtain the condensed features by calculating the importance of each sequence block. 
Then, we add the condensed features to the sampled features and use short text features to sample and fuse long video features.
We share the self-attention and reserve different feed forward networks for different modalities in the sampler. 

Figure~\ref{fig:intro} shows that previous models (a)\cite{Fu2021VIOLETE, Huang2022CloverTA} and (b)\cite{Li2021AlignAP} consume substantial video memory with their lengthy concatenated feature fusion. 
Both (b) and (c)\cite{Alayrac2022FlamingoAV} compress video features, a common choice due to their greater length compared to text. However, excessive compression can impair performance because of the rich information in video features. 
In contrast, we design MuLTI like (d) and introduces the Text-Guided MultiWay-Sampler to efficiently condense text features for fusion. Since text is more concise, we use the streamlined text to direct video feature sampling, resulting in enhanced performance.

Pretraining on large-scale video-text datasets could improve the performance of video-text models significantly.
However, there are still domain gaps between the existing pretraining tasks and downstream tasks, specifically in videoQA. 
The difficulty of introducing videoQA to pretraining tasks is constructing suitable question-answer pairs. 
To reduce the domain gap between the pretraining task and the downstream task in videoQA, we introduce a new pretraining task named Multiple Choice Modeling (MCM). 
The MCM can bridge the task gap between pretraining and downstream tasks by constructing multiple-choice question answering tasks on large-scale video-text datasets.
It asks the model to find text descriptions that match the video most from a randomly constructed collection, which enhances the representation ability of the video and text encoders and the alignment between video and text features.

The contributions can be summarized as follows:~

\textbf{(1)}~
We propose MuLTI, a highly accurate and memory-efficient video-and-language framework, which achieves efficient and effective feature fusion through the feature sampling and attention modules.

\textbf{(2)}~
We propose a Text-Guided MultiWay-Sampler to sample long sequence features and facilitate the interactions between video and text features, reducing memory cost and improving performance.  


\textbf{(3)}~
We design a new pretraining task called Multiple Choice Modeling (MCM) to bridge the task gap between pretraining and downstream tasks. Experimental results on seven English tasks and one Chinese multi-label classification task demonstrate the effectiveness of MuLTI.

Although we designed MuLTI for industrial scenarios with long sequences, MuLTI still handles short sequences well and achieves state-of-the-art performance.


\section{Related Work}\label{sec:relatedwork}
%
\begin{figure*}[ht]
\centering
\includegraphics[width=0.99\textwidth]{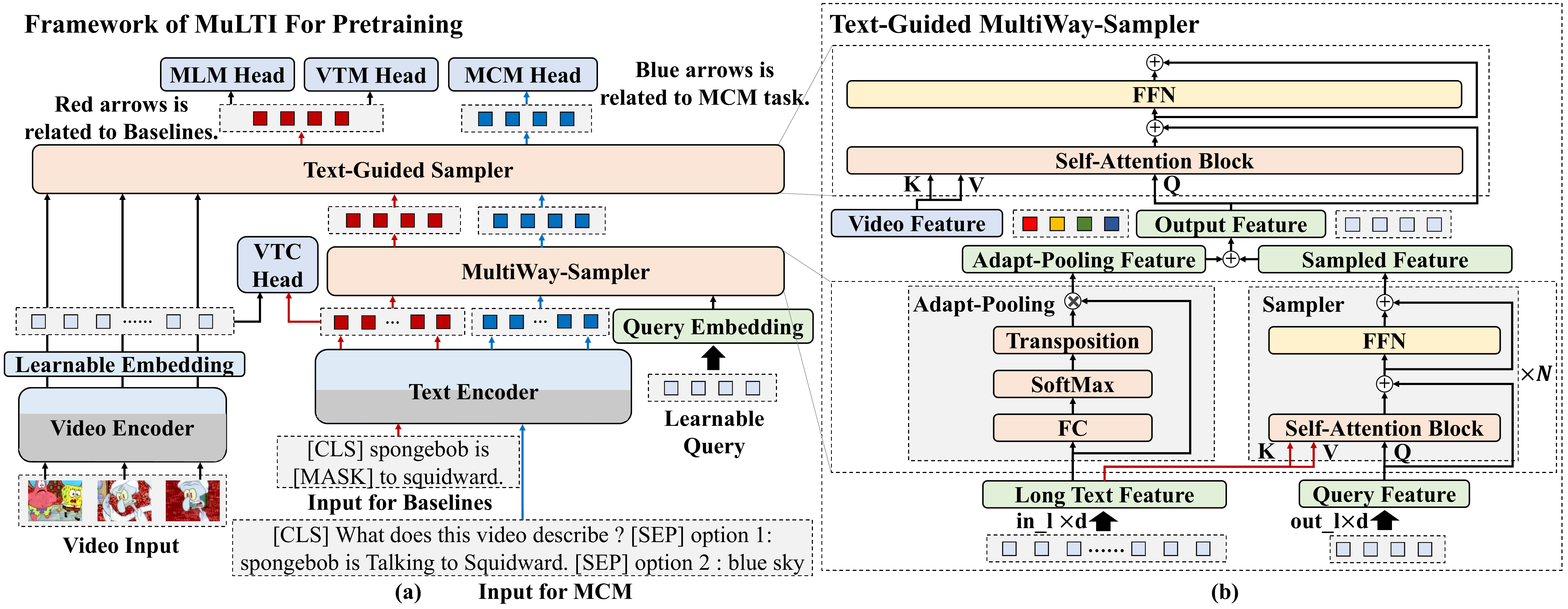}
\caption{
    (a) shows the framework of MuLTI. MuLTI contains a video encoder, a text encoder, and a Text-Guided MultiWay-Sampler. Text-Guided MultiWay-Sampler is used to condense the extracted features and feature fusion. 
    (b) shows the framework of the Text-Guided MultiWay-Sampler. The adapt-pooling feature provides origin information. We share the self-attention module and reserve different feed forward networks for different modalities in the sampler to accommodate modalities. 
}
\label{fig:arch}
\end{figure*}

\textbf{Video-and-Language Structure.} Clover~\cite{Huang2022CloverTA} and VIOLET~\cite{Fu2021VIOLETE} directly concatenate video and text features, using an encoder to manage their complex interactions, with complexity tied to the concatenated sequence length squared. 
ALPRO~\cite{Li2021AlignAP} similarly uses an encoder for fusing features but applies mean pooling on video features before concatenation, risking loss of crucial details. 
AllInOne~\cite{Wang2022AllIO} reduces memory demands by merging text with image features frame-by-frame but still faces high computational loads with extensive OCR transcripts.
Flamingo \cite{Alayrac2022FlamingoAV} attempts cost-cutting by condensing video features using samplers and random queries, which isn't ideal.
To tackle the above problems, we design a Text-Guided MultiWay-Sampler based on adapt-pooling residual mapping and self-attention modules to sample long sequence features and fuse multi-modal features.

\textbf{Video-and-Language Pretraining.} Four typical pretraining tasks are applied in previous pretraining framework: Masked Frame Modeling (MVM) tasks~\cite{Lei2021UnderstandingCV, Ma2021Top1SO, Fu2021VIOLETE,Huang2022CloverTA}, Masked Language Modeling (MLM) tasks~\cite{Devlin2018Bert,Sun2019VideoBERTAJ,Zhu2020ActBERTLG,Luo2020Univl,Li2020HeroHE,Lei2021Less,Fu2021VIOLETE}, Video Text Matching (VTM) and Video Text Comparison (VTC) tasks~\cite{Li2020HeroHE,Luo2020Univl,Fu2021VIOLETE,Li2021AlignAP}. MVM is used for video encoder optimization, MLM is used for text encoder optimization, VTM and VTC are used for joint optimization of video and text encoders.
%
%
In \cite{Ge2022BridgeFormerBV}, Multiple Choice Questions (MCQ) is proposed to learn fine-grained video and text features. 
However, MCQ is trained by contrastive loss and does not correlate well with videoQA. 
In summary, downstream task gaps persist between pretraining and downstream tasks, particularly in videoQA.
To address this, we enhance MuLTI with MCM, bridging the pretraining and downstream tasks.
\section{Methodology}

%
%

\subsection{MuLTI’s Architecture}\label{sec:method-arch}
Figure~\ref{fig:arch}~(a) gives an overview of MuLTI's architecture.
%
Details for each component are as follows.

\textbf{Video and Text Encoders:}~
Unless specified, a 12-layer VIT-B/16$_{224}$~\cite{Radford2021LearningTV} is used as video encoder. 
We sparsely sample $N_v$ frames from the input video. The VIT-B/16$_{224}$ model divides each frame into $K$ non-overlapping patches. 
The per-video features is $\Tilde{\Vec{v}}\in\mathbb{R}^{N_v\times K\times d}$, where $d$ is the feature dimension.
The output of the video encoder is a video features sequence: $\{\Vec{v}_1,...,\Vec{v}_{N_v}\}$, with $\Vec{v}_i\in\mathbb{R}^{K\times d}$. 
Experiments revealed the class token is unnecessary and was removed to save computation.
Unless specified, a 12-layer bert~\cite{Devlin2018Bert} is used as the text encoder. Assuming the length of input text is $N_t$, the output of the text encoder is a text features sequence $\Tilde{\Vec{t}}\in\mathbb{R}^{N_t\times d}$: $\{\Vec{t}_\mathrm{cls}, \Vec{t}_1,...,\Vec{t}_{N_t}\}$, with $\Vec{t}_i\in\mathbb{R}^{d}$. The $\Vec{t}_\mathrm{cls}$ is the output of the text \texttt{[CLS]} token. 
Following \cite{Miech2019HowTo100MLA,Sun2019VideoBERTAJ,Li2020HeroHE}, we explore training strategies for partially freezing encoder layers.

\textbf{Text-Guided MultiWay-Sampler:}~The multi-modal fusion core is the Text-Guided MultiWay-Sampler, adapted from the transformer decoder, shown in Figure~\ref{fig:arch}~(b). The Text-Guided MultiWay-Sampler is designed to condense text features and fuse different modal features efficiently. 
Following \cite{Alayrac2022FlamingoAV}, we initialize a random learnable query to condense features via sampling.
%
%
The expression $Sampler(z, q)$ represents the sampling of feature $z$ using the query vector $q$ through the sampler. 

(i) \textbf{Why we need Adapt-Pooling?} Learnable queries can compress features well, but starting with random vectors may reduce their effectiveness.
Random initialization may lose key details in original features, weakening the model's ability to capture and retain the essence of the data.
Therefore, we design an attention-based lightweight Adapt-Pooling method to condense long sequence features. 
The Adapt-Pooling structure is shown on the left side of the Figure~\ref{fig:arch}~(b). 
The formula is shown below, $AdaPool(z)$ is the output of the Adapt-Pooling, with $W^{reduce} \in \mathbb{R}^{d\times N_s}$, $d$ the hidden dimension of the transformer, $N_s$ the length of condensed features, $*.T$ the transposition of the matrix. 

\begin{equation}
AdaPool(z) = Softmax((W^{reduce}z).T) \times z
\end{equation}

The $Softmax((W^{reduce}z).T)$ yields an importance weight matrix of shape $[N_s, N_i]$, with each element signifying the relative importance of the corresponding block within the sequence, and $N_i$ representing the length of the input features. 
Adapt-Pooling selectively highlights key input segments, condensing features while preserving its critical attributes. 
This integration enriches the feature set with distilled information and ensures full data utilization, boosting the model's capacity and robustness. 

(ii) \textbf{Why we condense text features?} 
The video features are often redundant, whereas text features are denser and more meaningful~\cite{He2022MaskedAA}. Language guidance is key to distilling valuable information from videos.
Both \cite{Li2021AlignAP} and \cite{Alayrac2022FlamingoAV} condense the video features. Excessive compression harms model performance; using condensed text to sample video features improves results. 
Before fusion, learnable time embeddings enhance image features for temporal modeling. The short text features are used to sample the video features to fuse multi-modal features. 
In our Text-Guided MultiWay-Sampler, we use shared self-attention modules but distinct FFNs for each modality to handle multi-modal features efficiently.
The fuse feature is shown as follow, with $q$ the query embedding of text features, $z_{out}$ the fused feature:
\begin{equation}
z_{out} = Sampler(\Tilde{\Vec{v}}, Sampler(\Tilde{\Vec{t}}, q) + AdaPool(\Tilde{\Vec{t}}))
\end{equation}

A work similar to ours is Token Learner~\cite{Ryoo2021TokenLearnerWC}, which uses spatial attention in model to extract 8 or 16 representative vectors from an image.
The difference is that we use Adapt-Pooling and self-attention to condense features for multi-modal fusion. The sampler extracts complex information via self-attention, while Adapt-Pooling provides fast, simple features through residual mapping.

(iii) \textbf{Why Text-Guided MultiWay-Sampler is efficient ?} Our feature fusion module outperforms flatten-based methods and transformer encoders in efficiency, as simple analysis shows: we assume VIT-B/16$_{224}$ is used as video encoder, each frame will be flattened into a sequence of length 196. Let the number of queries be $N^q_t$ for text, the length of the video features be $N_v \times 196$, and the length of the text features be $N_t$. Thus the complexity of the flatten method will be $O((N_t+N_v \times 196)^2)$. After applying the Text-Guided MultiWay-Sampler, the complexity is $O(N^q_t \times N_v \times 196 + N^q_t \times N_t)$. As $N^q_t$ are generally much smaller than $N_v$ and $N_t$. our method is much more efficient than other methods. 

\begin{table}[!t]
\small
\centering	
\begin{tabular}	{l | c c c}
    \toprule
        \textbf{Method} &
        \textbf{FLOPs} &
        \textbf{Params} &
        \textbf{FPS} \\
        \midrule	
        \textbf{MuLTI-S} & 99G & 203M & 20.74\\
        \textbf{MuLTI-B} & 346G & 247M & 10.13\\
        \textbf{VIOLET~\cite{Fu2021VIOLETE}} & 249G & 198M & 9.05\\
        \textbf{ALPRO~\cite{Li2021AlignAP}} & 432G & 235M & 9.97\\
        \midrule
        \textbf{MuLTI-L} & 1509G & 746M & 3.12\\ 
        \textbf{FrozenBiLM~\cite{Yang2022ZeroShotVQ}} & 1733G & 1224M & 2.54\\
    \bottomrule
\end{tabular}
\caption{
    Comparison among models with 16 frames. Text length is 512. FPS is based on 1 NVIDIA V100 16GB GPU.
}
\label{tbl:flops_params}
\end{table}		
\textbf{MuLTI for different scenes.}~In this section, we built scalable models for resource-varied scenes. We replace the video encoder from VIT-B/16 to VIT-L/14 and the text encoder from bert-base to bert-large. Then, we obtain MuLTI-L. 
In addition, we replace the video encoder from VIT-B/16 to VIT-B/32 and reduce the text encoder from 12 layers to 6 layers. Then, we obtain MuLTI-S. The floating point of operations~(FLOPs), parameters~(Params) and frames per second~(FPS) of different models are shown in Table~\ref{tbl:flops_params}.

\subsection{Pretraining for MuLTI}\label{sec:method-pre}

\textbf{Multiple Choice Modeling:}~
Despite MLM and VTM's success in learning video-text representations, a significant gap remains between pretraining and downstream tasks like videoQA.
The difficulty of introducing videoQA into the pretraining task is constructing suitable question-answer pairs.
Inspired by multiple choice videoQA, we find the text descriptions paired with videos are the correct natural answers.
Therefore, we introduce Multiple Choice Modeling, a new pretraining task that bridges the task gap between pretraining and downstream tasks. Specifically, it is constructed as follows, which is a four-choice question.

\noindent{\small\texttt{"[CLS]<Question> ? [SEP] Option 1: <Answer 1>. [SEP] Option 2: <Answer 2>. [SEP] Option 3: <Answer 3>. [SEP] Option 4: <Answer 4>."}}

We randomly place the correct descriptions in \texttt{<Answer 1>, <Answer 2>, <Answer 3>, <Answer 4>}, and obtain answers other than the correct descriptions through the text corpus. The \texttt{<Question>} also has various choices, such as ``What does this picture describe?'', ``What does this video describe?'' and so on. 

As shown in Figure~\ref{fig:arch}~(a), typical MLM, VTM, and VTC tasks correspond to the red arrows and red squares in the image. 
The MCM corresponds to the image's blue arrows and blue squares, and the MCM does not conflict with the other pretraining tasks. 
The MCM is seamlessly integrated with other pretraining tasks and does not require additional manual annotations or data preprocessing. 
It utilizes video encoders to extract visual features and text encoders for generating textual representations, followed by a Text-Guided MultiWay-Sampler for feature fusion. 
The MCM head evaluates the given options' relevance to the video, optimizing alignment using cross-entropy loss.
The MCM task, choosing the best description from options, mirrors essential videoQA cognition, enhancing the model's cross-modal reasoning and alignment.
MCM directly improves the model's ability to match text with corresponding videos, enhancing performance in text-video retrieval tasks. 

\textbf{Pretraining Objectives:}~We also employ the MLM, VTM and VTC, considering their effectiveness.  
The MLM randomly masks input tokens with 15\% probability and replaces them with [MASK], which are predicted based on video and text.
The VTC treats matching video text pairs as positive pairs and other video text pairs in the batch as negative pairs.
The VTM is slightly different from VTC, where the multi-modal features are fused before used for classification. 
The overall pretraining objective of MuLTI is:

\begin{equation}
L = L_{mlm} + L_{vtc} + L_{vtm} + L_{mcm}
\end{equation}

\begin{table*}[!]
\centering
\small
{
\centering	
    \begin{tabular}{ll|ccccc llll}
        \toprule
             ~ & ~ & \multirow{2}{*}{MSRQ} & \multirow{2}{*}{MSVQ} & TGIF. & TGIF. & TGIF & \multicolumn{2}{c}{MSRVTT}  & \multicolumn{2}{c}{DiDeMo} \\
             ~ & ~ & ~ & ~ & Act. & Tran. & Fra. & \multicolumn{2}{c}{Ret} & \multicolumn{2}{c}{Ret} \\
    
            Method & \#PT & Acc.$\uparrow$ & Acc.$\uparrow$ & Acc.$\uparrow$ & Acc.$\uparrow$ & Acc.$\uparrow$ & R1 / R5 / R10 $\uparrow$ & G-M $\uparrow$ & R1 / R5 / R10 $\uparrow$ & G-M $\uparrow$\\
    
            \midrule
            CLIP4CLIP & 400M & - & - & -  & -  & -  & 43.1 / 70.4 / 80.8 & 62.6 & 43.4 / 70.2 / 80.6 & 62.6 \\
            QB-Norm & 400M & - & - & -  & -  & -  & 47.2 / 73.0 / 83.0* & 65.9* & 43.3 / 71.4 / 80.8* & 63.0* \\
            CAMoE & 400M & - & - & -  & -  & -  & 47.3 / 74.2 / 84.5* & 66.7* & 43.8 / 71.4 / 79.9* & 63.0* \\  
            TS2-Net & 400M & - & - & -  & -  & -  & 54.0 / 79.3 / 87.4* & 72.1* & 47.4 / 74.1 / 82.4* & 66.1* \\
            ALPRO & 5.5M & 42.1 & 45.9 & -  & -  & -  & 33.9 / 60.7 / 73.2 & 53.2 & 35.9 / 67.5 / 78.8 & 57.6 \\
            VIOLET & 185.5M & 43.9 & 47.9 & 92.5 & 95.7 & 68.9 & 34.5 / 63.0 / 73.4 & 54.2 & 32.6 / 62.8 / 74.7 & 53.5\\
            AllInOne & 102.5M & 44.3 & 47.9 & 92.7 & 94.3 & 64.2 & 37.9 / 68.1 / 77.1  & 58.4  & 32.7 / 61.4 / 73.5 & 52.8  \\
            Clover & 5.5M & 44.1 & 52.4 & 95.0 & 98.2 & 71.6 & 40.5 / 69.8 / 79.4 & 60.7 & 50.1 / 76.7 / 85.6 & 69.0 \\
            Flamingo & 2139M & 47.4 & 52.3 & -  & -  & - & - & - & - & -  \\ 
            FrozenBiLM & 10M & 47.0 & 54.4 & -  & -  & 68.6 & - & - & - & -  \\ 
            \midrule 
            \multirow{2}{*}{\textbf{MuLTI-S}} & \multirow{2}{*}{5.5M} & \multirow{2}{*}{45.6} & \multirow{2}{*}{50.0} & \multirow{2}{*}{97.3} & \multirow{2}{*}{98.9}  & \multirow{2}{*}{71.2} & 41.3 / 70.6 / 79.7 & 61.5 & 42.6 / 71.4 / 80.0 & 62.5 \\
             & & & & & & & 45.8 / 73.5 / 82.0* & 65.1* & 47.9 / 73.0 / 82.6* & 66.1* \\ 
            \midrule
            \multirow{2}{*}{\textbf{MuLTI-B}} & \multirow{2}{*}{5.5M} & \multirow{2}{*}{46.6} & \multirow{2}{*}{53.0} &  \multirow{2}{*}{97.3} & \multirow{2}{*}{\textbf{99.1}} & \multirow{2}{*}{73.5} & 45.1 / 72.4 / 81.8  & 64.4 & 45.2 / 74.6 / 82.2 & 65.2 \\
             & & & & & & & 49.4 / 75.9 / 84.0* & 68.0* & 48.3 / 75.4 / 83.5* & 67.2* \\ 
            \midrule
            \multirow{2}{*}{\textbf{MuLTI-L}} & \multirow{2}{*}{5.5M} & \multirow{2}{*}{\textbf{47.8}} & \multirow{2}{*}{\textbf{54.7}} & \multirow{2}{*}{\textbf{97.9}} & \multirow{2}{*}{99.0} & \multirow{2}{*}{\textbf{75.6}} & 48.7 / 75.9 / 83.9 & 67.7  & 50.5 / 78.5 / 86.2 & 69.9 \\
             & & & & & & & \textbf{54.7} / \textbf{77.7} / \textbf{86.0*} & \textbf{71.5*} & \textbf{56.5} / \textbf{80.2} / \textbf{87.0*} & \textbf{73.3*} \\ 
        \bottomrule
    \end{tabular}
}
\caption
{
     Comparisons with existing methods. \#PT means number of pretrain datasets. Acc. (\%) denotes the performance of videoQA. R@k denotes recall (\%) with k retrieval efforts. G-M denotes the geometric mean of R@1, R@5, R@10. The datasets commonly used are
     WebVid2M~\cite{Bain2021FrozenIT}, WebVid10M~\cite{Bain2021FrozenIT}, WIT~\cite{Radford2021LearningTV}, HowTo100M~\cite{Miech2019HowTo100MLA},  
     YT-Temporal-180M~\cite{Zellers2021MERLOTMN}, Conceptual Captions~\cite{Sharma2018ConceptualCA}. * indicates that  DSL~\cite{Cheng2021ImprovingVR} or QB-Norm~\cite{Bogolin2021CrossMR} is used for post-processing. 
}

\label{tbl:main}
\end{table*}

\section{Experiments}

%

\subsection{Implementation Details}\label{sec:method-impl}

\textbf{Pretraining datasets:}~We pretrained the model using two large datasets. One is WebVid-2M~\cite{Bain2021FrozenIT}, which contains 2.5M video-text pairs. Because pretraining the video-text model using image-text pairs also improves the model's performance~\cite{Lei2021Less}, the CC-3M\cite{Sharma2018ConceptualCA} is also used as a pretrained dataset containing 3M image-text pairs. 

\begin{table}[!t]
\centering
\small
\begin{tabular}	{l c c | c}
    \toprule
    \textbf{Method} & \textbf{\#PT} & \textbf{OCR} & \textbf{Multi-Label} \\
    \midrule
    VIOLET $\ddag$ & - & \ding{56} & 55.22\\
    ALPRO $\ddag$ & - & \ding{56} & 58.53\\
    \midrule
    \textbf{MuLTI-S} &  - & \ding{56} & 63.97\\
    \textbf{MuLTI-S} & -  &  \ding{52} & \textbf{66.13}\\
    \midrule
    \textbf{MuLTI-B} &  - & \ding{56} & 64.60\\
    \textbf{MuLTI-B} & - &  \ding{52} & \textbf{67.86}\\
    \bottomrule
\end{tabular}
\caption
{
    Comparisons on multi-label classification in mAP (\%). $\ddag$ means the methods are reproduced in our framework.
}
\label{tbl:multi-label}
\end{table}
\begin{figure}[!t]
\centering
\includegraphics[width=0.465\textwidth]{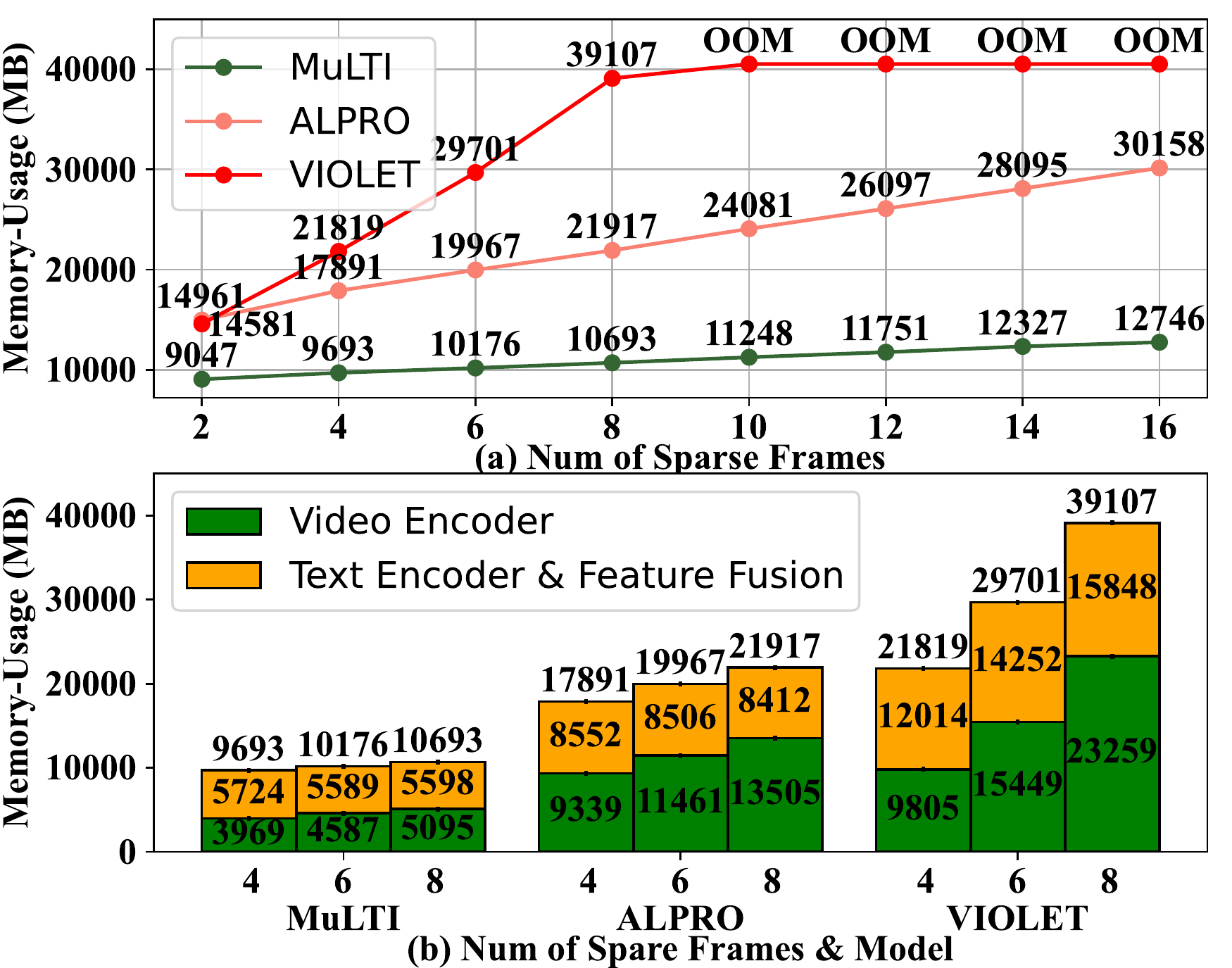}
\caption{
Comparisons with existing methods on Memory-Usage with different numbers of frames. Text length is 512. 
}
\label{fig:memory_models}
\end{figure}

\textbf{We implement MuLTI in PyTorch~\cite{paszke2019pytorch}.} In detail, the video encoder is initialized with pretrained weights from CLIP~\cite{Radford2021LearningTV}. Text encoder is initialized with a 12-layer of the BERT$_{\mathrm{base}}$ model~\cite{Devlin2018Bert}. 
Then, a 4-layer Text-Guided MultiWay-Sampler is used to condense text features and fuse multi-modal features.
The length of query embedding is set to 16. 
MuLTI pretraining spanned 10 epochs on eight NVIDIA A100 GPUs, a 256 batch size totaling 200k iterations. 
Optimization used AdamW with a $1e^{-4}$ learning rate, 0.05 weight decay, and a warm-up scheduler.
We uniformly sample 16 frames for each video and scale them to $224\times 224$.

\subsection{Downstream Tasks and Datasets}\label{sec:exp-setup}

\textbf{Video Question Answering.} We evaluate MuLTI on five widely used videoQA tasks.
\textbf{(1)}~\textbf{MSRQ (MSRVTT-QA)}~\cite{xu2017video, Xu2016Msr} is a open-ended videoQA task includes 10k videos and 243k question-answer pairs.
\textbf{(2)}~\textbf{MSVQ (MSVD-QA)}~\cite{xu2017video, chen2011collecting} is a open-ended videoQA task includes 1970 videos and 50k question-answer pairs.
\textbf{(3)}~\textbf{TGIF-QA \cite{Jang2017TGIFQATS}} contains three datasets: TGIF-Action and TGIF-Transition for multiple-choice videoQA tasks, and TGIF-Frame for open-ended videoQA tasks.

\textbf{Text-Video Retrieval.} \textbf{(1)}~\textbf{MSRR (MSRVTT-Ret)} contains 10K videos with 200K annotations. Following \cite{Fu2021VIOLETE}, we use 9k videos for training and 1k videos for testing.
\textbf{(2)}~\textbf{DiDeMo (DiDeMo-Ret)} consists of 10K videos with 40K annotations. Following \cite{Lei2021Less}, we concatenate all annotations from the video into a title. 

\begin{table*}[htb]
\small
\centering
\begin{tabular}	{l | c | c c c | c c c c c c}
    \toprule 
    \multirow{2}{*}{\textbf{Method}} &
    \multirow{2}{*}{\textbf{Base}}& 
    \multirow{2}{*}{\textbf{TGMS}} &  
    \multirow{2}{*}{\textbf{PB}} &  
    \multirow{2}{*}{\textbf{MCM}} &  
    \textbf{MSRQ} & 
    \textbf{MSVQ} &
    \multicolumn{4}{c}{\textbf{MSRR}} \\
    &  &  &  &  & \footnotesize\textbf{Acc.}$\uparrow$ & \footnotesize\textbf{Acc.}$\uparrow$ & \footnotesize \textbf{R1}$\uparrow$ & \footnotesize\textbf{R5}$\uparrow$
    & \footnotesize\textbf{R10}$\uparrow$ & \footnotesize\textbf{G-Mean} $\uparrow$ \\
    \midrule
    \multirow{4}* {MuLTI-B} & \ding{52} & \ding{56} & \ding{56} & \ding{56} & 44.84 & 48.35 & 38.90 & 69.50 & 78.50 & 59.64 \\
    & \ding{52} & \ding{52} & \ding{56} & \ding{56} & 45.54 & 49.86 & 38.80 & 70.30 & 80.10 & 60.22 \\
    & \ding{52} & \ding{52} & \ding{52} & \ding{56} & 46.28 & 51.93 & 44.30 & 72.40 & \textbf{81.90} & 64.04 \\
    & \ding{52} & \ding{52} & \ding{52} & \ding{52} & \textbf{46.61} & \textbf{53.03} & \textbf{45.10} & \textbf{72.40} & 81.80 & \textbf{64.40} \\
    \bottomrule
    \end{tabular}

\caption
{Evaluations of the proposed methods. TGMS: Text-Guided MultiWay-Sampler. PB (Pretraining Baseline): Pretraining model with MLM, VTM and VTC. MCM: Multiple Choice Modeling. Acc. (\%) is used to measure the performance of videoQA. R@k denotes recall (\%) with k retrieval efforts. G-Mean denotes the geometric mean of R@1, R@5, R@10.}
\label{tbl:main_proposal}
\end{table*}		
\begin{figure}[!t]
\centering
\includegraphics[width=0.47\textwidth]{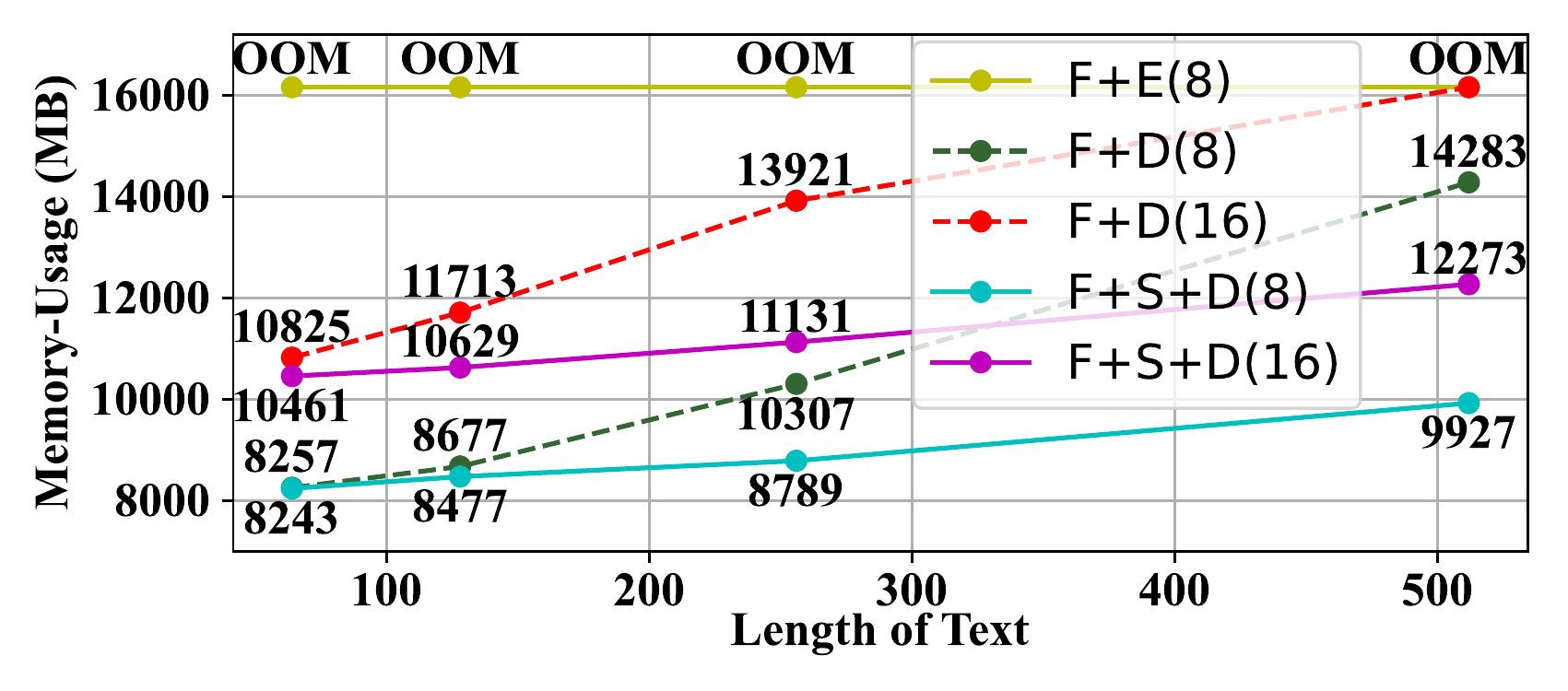}
\caption{
Comparisons of different text length and number of frames on Memory-Usage. The F means Flatten, the D means Decoder, the E means Encoder, the S means Sampler. The number in parentheses represents the number of frames. 
}
\label{fig:memory_sampler}
\end{figure}
\begin{figure}[!t]
\centering
\includegraphics[width=0.45\textwidth]{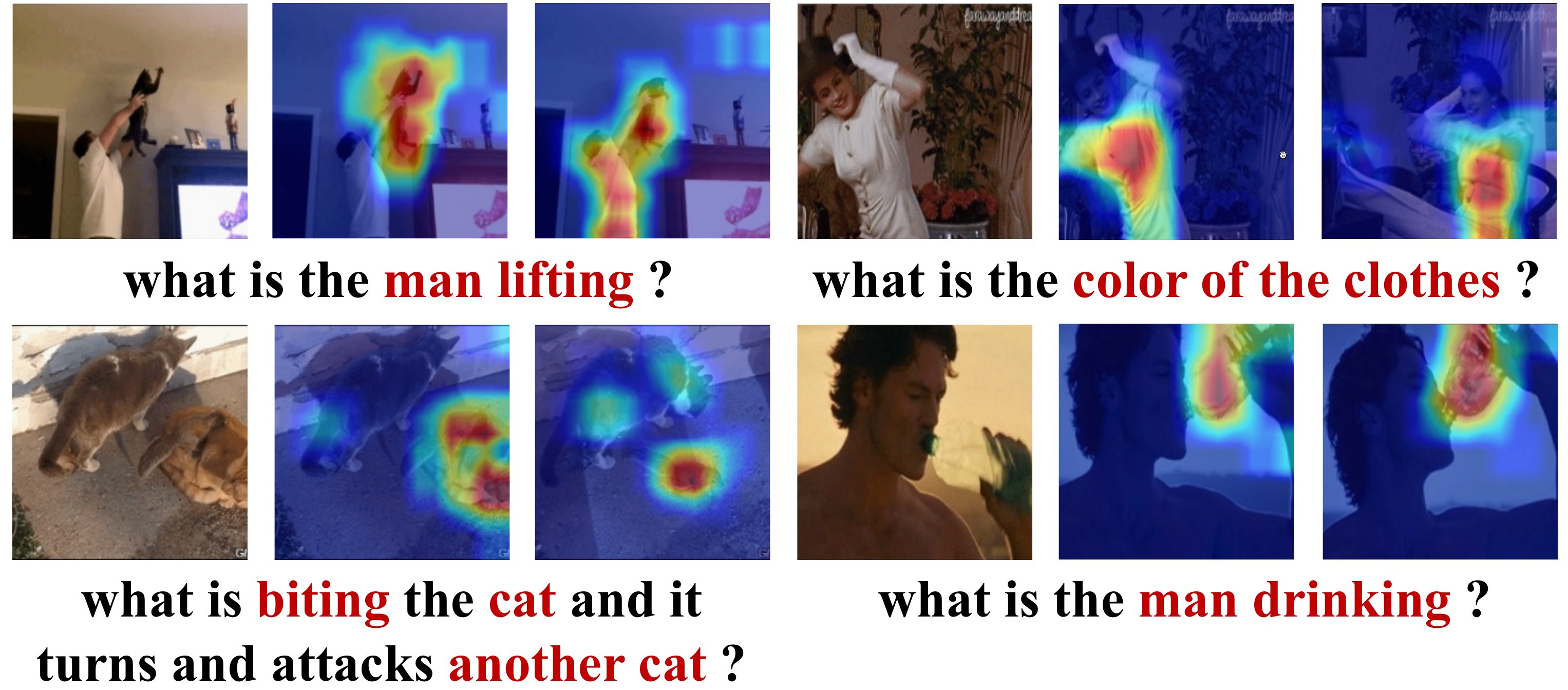}
\caption{
A visualization of the cross-attention map from the Text-Guided MultiWay-Sampler. 
}
\label{fig:example_att}
\end{figure}

\textbf{Multi-Label Classification.}
Video labels are crucial for the ranking models used in online advertising\footnote{https://algo.qq.com/index.html}. 
We create a short video dataset from our app, which includes
486k videos with captions and 21696 labels. Multiple professional editors cross check the labels. We used a high-performing text detector from ICDAR\footnote{https://rrc.cvc.uab.es/?ch=4\&com=evaluation\&task=4} for OCR transcripts. The OCR transcripts are truncated to 512. The examples for multi-label classification can be found in the appendix.

\subsection{Performance of Proposed Methods}\label{sec:exp-proposes-method}

Table~\ref{tbl:main} compares MuLTI with CLIP4CLIP~\cite{Luo2021CLIP4ClipAE}, QB-Norm~\cite{Bogolin2021CrossMR}, CAMoE~\cite{Cheng2021ImprovingVR}, TS2-Net~\cite{Liu2022TS2NetTS}, ALPRO~\cite{Li2021AlignAP}, VIOLET~\cite{Fu2021VIOLETE}, AllInOne~\cite{Wang2022AllIO}, Clover~\cite{Huang2022CloverTA}, Flamingo ~\cite{Alayrac2022FlamingoAV} and FrozenBiLM~\cite{Yang2022ZeroShotVQ}.

In videoQA tasks, MuLTI surpasses all baseline models on MSRQ, MSVQ, TGIF-Action, TGIF-Transition and TGIF-Frames. 
Since MuTLI does not use speech data as input, it is compared with FrozenBiLM\cite{Yang2022ZeroShotVQ} without using speech data. In general, MuLTI achieves state-of-the-art performance in various QA tasks. 

In text-video retrieval tasks, we finetune MuLTI using the MSRVTT and DiDeMo datasets. Our results demonstrate that MuLTI is highly competitive in both benchmarks, particularly in the DiDeMo dataset. These findings highlight the effectiveness of MuLTI for text-video retrieval.

\begin{table}[!t]
\centering
\small
\begin{tabular}	{l | c c c c}
    \toprule
    \textbf{Methods} & \textbf{MSRQ} & \textbf{MSVQ} & \textbf{Memory Usage} \\
    \midrule
        \textbf{Class Token} & 44.54 & 47.90 & 7081 \\
        \textbf{Mean Pooling} & 44.40 & 47.07 & 6941 \\
        \textbf{Max Pooling} & 44.41 & 46.93 & 6963 \\
        \textbf{Flatten + Encoder} & 44.84 & 48.35 & 15791 \\
    \midrule
        \textbf{TGMS} & \textbf{45.54} & \textbf{49.86} & 10551 \\
    \bottomrule
\end{tabular}
\caption
{
Ablation studies on feature retention methods. The number of sparse frames is set to 6 for Flatten method. TGMS means Text-Guided MultiWay-Sample.
}
\label{tbl:cls2flatten}
\end{table}
\begin{table}[!t]
\small
\centering
\begin{tabular}	{l | c c c c |c c}
    \toprule
        \textbf{Method} & \textbf{CV} & \textbf{CT} & \textbf{SS} & \textbf{AP} & \textbf{MSRQ} & \textbf{MSVQ}\\
    \midrule
         & \ding{56} & \ding{56} & \ding{56} & \ding{56} & 45.13 & 49.19 \\
         & \ding{52} & \ding{56} & \ding{56} & \ding{56} & 44.76 & 48.10  \\
         & \ding{52} & \ding{56} & \ding{52} & \ding{52} & 45.14 & 48.92  \\
         \textbf{Flatten} & \ding{52} & \ding{52} & \ding{56} & \ding{56} & 44.57 & 48.50 \\
         \textbf{Decoder} & \ding{56} & \ding{52} & \ding{56} & \ding{56} & 45.08 & 49.38  \\
         & \ding{56} & \ding{52} & \ding{52} & \ding{56} & 45.16 & 49.80  \\
         & \ding{56} & \ding{52} & \ding{56} & \ding{52} & 45.48 & 49.54  \\
         & \ding{56} & \ding{52} & \ding{52} & \ding{52} & \textbf{45.54} & \textbf{49.86}  \\
    \bottomrule
\end{tabular}
\caption{An ablation study on feature compression methods. CV means Condensed Video, CT means Condensed Text, SS means Shared-Sampler, AP means Adapt-Pooling. }
\label{tbl:sampler_decoder}
\end{table}
For multi-label classification, we compare MuLTI with VIOLET and ALPRO but exclude FrozenBiLM due to its impractical size for industry deployment.
VIOLET and ALPRO do not use OCR transcripts as they would lead to out-of-memory on V100 GPUs. We also report MuLTI's OCR-less performance in Table~\ref{tbl:multi-label} for a fair comparison; MuLTI significantly surpasses both VIOLET and ALPRO.
As shown in Figure~\ref{fig:memory_models}, MuLTI maintains a video memory cost less than half of ALPRO's and VIOLET's when frame count rises during training, because its efficient fusion modules minimizes memory cost increases.

\begin{table}[!t]
\small
\centering
\begin{tabular}	{ c c c | c c c }
    \toprule 
    \textbf{PB} & \textbf{MVM} & \textbf{MCM} & \textbf{MSRQ} & \textbf{MSVQ} & \textbf{MSRR}  \\ 
    \midrule
     \ding{52} & \ding{56} & \ding{56} & 46.28 & 51.93 & 64.04 \\
     \ding{52} & \ding{52} & \ding{56} & 45.87 & 50.16 & 63.41 \\
     \ding{52} & \ding{52} & \ding{52} & 46.11 & 51.65 & 63.71 \\
     \ding{52} & \ding{56} & \ding{52} & \textbf{46.61} & \textbf{53.03} & \textbf{64.40} \\
    \bottomrule
\end{tabular}
\caption
{
Ablation studies on the Multiple Choice Modeling. PB means Pretraining Baseline. 
}
\label{tbl:mcm}
\end{table}		
\begin{table}[!t]
\centering
\small
\begin{tabular}	{c c | c c c}
    \toprule 
     \multicolumn{2}{c|}{\textbf{Frozen / Total}} & \multirow{2}{*}{\textbf{MSRQ}} & \multirow{2}{*}{\textbf{MSVQ}} & \multirow{2}{*}{\textbf{Memory Usage}}  \\ 
    \specialrule{0em}{1pt}{0pt} \cline{1-2} \specialrule{0em}{1pt}{0pt}
     \textbf{VE} & \textbf{TE} &  &  &  \\
    \midrule
     12/12 & 12/12 & 44.06 & 46.83 & 6109  \\
     12/12 & 0/12 & 44.07 & 47.12 & 7439 \\
     6/12 & 0/12 & 45.10 & 47.57 & 18219 \\ 
     9/12 & 0/12 & 45.59 & 47.52 & 11541 \\
     9/12 & 3/12 & 45.50 & 49.63 & 11131 \\
     9/12 & 6/12 & \textbf{45.54} & \textbf{49.86} & 10551 \\
     9/12 & 9/12 & 45.04 & 49.14 & 10283 \\
    \bottomrule
\end{tabular}
\caption
{
Ablation studies on frozen layers.
VE refers to video encoder and TE refers to text encoder. 
Frozen/Total refers to the number of frozen and total layers. 
}
\label{tbl:partially-frozen}
\end{table}		
\begin{table}[!t]
\small
\centering
\begin{tabular}	{c c | c c c c}
    \toprule 
    \textbf{Adapter} & \textbf{ATT} & \textbf{MSRQ} & \textbf{MSVQ} & \textbf{MSRR}  & \textbf{DiDeMo} \\ 
    \midrule
     \ding{56} & \ding{56} & 45.54 & 49.86 & 60.22 & 51.68\\
     \ding{52} & \ding{56} & 45.61 & 50.48 & 60.54 & 52.08 \\
     \ding{52} & \ding{52} & \textbf{45.71} & \textbf{50.63} & \textbf{61.16} & \textbf{52.42} \\
    \bottomrule
\end{tabular}
\caption
{
Ablation studies on the Attention-Adapter. ATT means Attention.
}
\label{tbl:adapter}
\end{table}		

Finally, we evaluate our main technical contributions in Table~\ref{tbl:main_proposal}.
Compared with baseline models, our main technical contributions improve performance on all datasets. 
The Text-Guided MultiWay-Sampler boosts MuLTI's multi-modal fusion ability, pinpointing key details in surplus video features. 
MCM advances the model's alignment ability and narrows the gap between pretraining and downstream tasks.

\subsection{The Importance of Text-Guided MultiWay-Sampler}\label{sec:exp-exclusive}


\textbf{Why we condense text features?}~
%
We compare performance of different aggregation methods (\textit{i.e.} Class Token, Mean Pooling, Max Pooling and Flatten) in Table \ref{tbl:cls2flatten}. 
Results show that Flatten outperforms other aggregation methods but requires substantial video memory.
Above section reveals the decoder uses less memory than the encoder for long sequences, prompting its use in feature fusion. 
The decoder handles datasets like MSRQ well. However, the cost is still high when processing long text and video like our multi-label datasets. The specific memory cost is shown in Figure~\ref{fig:memory_sampler}. 
Following \cite{Alayrac2022FlamingoAV}, we use a decoder-based sampler for feature condensation
%
%
%
Table~\ref{tbl:sampler_decoder} compares different condensation methods, showing text compression's superiority.
As shown in Figure~\ref{fig:example_att}, the visual part most relevant to the problem is given more weight.

\textbf{The importance of Shared-Sampler.}~ 
The sampler and feature fusion module, using the same decoder structure, can share weights without compromising performance, simplifying model optimization~\cite{Wang2022ImageAA}.
We share the sampler and decoder's self-attention but keep separate FFNs for each modality, cutting parameters while maintaining performance.
Compared with the Flatten Method, the Shared-Sampler improves accuracies on MSRQ and MSVQ by 0.32\% and 1.45\%, respectively.

\textbf{The importance of Adapt-Pooling.}~
As shown in Table~\ref{tbl:sampler_decoder}, the sampler leads to worse performance when condensing text and video features. 
The sampler's random query vector carries the risk of losing original key features; we design a lightweight aggregation module, Adapt Pooling, to preserve the original features.
As shown in Table~\ref{tbl:sampler_decoder}, the Adapt-Pooling improves accuracy on MSRQ and MSVQ. 
Additionally, we explored various combination methods (\textit{i.e.} add, concatenate, and multiply), and noted slight performance differences.
We achieved an accuracy of 45.51\% using concatenate and 45.45\% using multiply on MSRQ.

To verify these techniques' robustness, we applied them to condense video features, which also improved performance.

\subsection{The Importance of Multiple Choice Modeling}\label{sec:exp-adapter}

%
%
MCM aims to bridge the gap between pretraining and downstream tasks by integrating videoQA into pretraining, enhancing the model's focus on video and sentence subjects for better multimodal feature extraction.

We use the classical MLM, VTM, and VTC tasks to pretrain the model as a baseline. Due to video content corruption caused by MVM, the MVM task conflicts with other tasks~\cite{Lei2021UnderstandingCV}. 
In our initial attempts to include MVM for pretraining, we observed a degradation in performance as shown in Table \ref{tbl:mcm}. Thus, we have decided not to use MVM for pretraining. 
To confirm MCM's robustness, we also added MCM for pretraining based on the usage of MVM. The results show MCM still substantially enhances model's performance.
Compared to the model pretrained with baseline, MCM explicitly improves the model's performance on the videoQA task by narrowing the task gap between pretraining and downstream tasks.
MCM's promotion of multi-modal feature alignment enhances the model's retrieval task performance.
As shown in Table \ref{tbl:mcm}, the models pretrained with MCM outperformed the baseline in both videoQA and retrieval tasks, demonstrating its effectiveness.

\subsection{Ablation Experiment on Training Strategies}\label{sec:exp-pfe}

\textbf{Analysis of Frozen Layers.} 
In this section, we systematically evaluate the effect of the number of frozen layers. The results on videoQA are demonstrated in Table~\ref{tbl:partially-frozen}. It indicates that unfreezing the top layers of video and text encoders can improve performance on both datasets. 

\textbf{Analysis of Attention-Adapter.} 
Analyzing frozen layers reveals that unfreezing excess layers reduces accuracy due to overfitting from excessive parameter adjustments.
Following \cite{Yang2022ZeroShotVQ}, we add adapters to the encoders in the shallow layers.
Table~\ref{tbl:adapter} shows that while adapters perform effectively, their capability is constrained by the basic FFN module. By integrating a lightweight attention module \cite{Hu2017SqueezeandExcitationN}, the model focuses better on informative tokens.

\section{Conclusion}
\label{sec:conclusion}
We present MuLTI, a high-performing video-language framework with a novel Text-Guided MultiWay-Sampler for improved sampling efficiency and a pretraining task to better align with downstream tasks. MuLTI achieves state-of-the-art performance on seven video-language benchmarks. 

\bibliography{aaai24}

\end{document}